\documentclass{svproc}

\usepackage{url}

\usepackage{multicol}
\usepackage{multirow}
\usepackage[bookmarks=true]{hyperref}

\usepackage{amssymb}
\usepackage{amsmath}
\usepackage{mathtools}
\usepackage{bm}

\usepackage{textcomp}

\usepackage[misc]{ifsym}

\usepackage{graphicx}
\usepackage{subcaption}

\usepackage{algpseudocode}
\usepackage{algorithm}

\usepackage{tikz}
\usetikzlibrary{decorations.pathreplacing,calc}
\newcommand{\tikzmark}[1]{\tikz[overlay,remember picture] \node (#1) {};}

\newcommand*{\AddNote}[4]{%
    \begin{tikzpicture}[overlay, remember picture]
        \draw [decoration={brace,amplitude=0.5em},decorate,ultra thick,black]
            ($(#3)!(#1.north)!($(#3)-(0,1)$)$) --  
            ($(#3)!(#2.south)!($(#3)-(0,1)$)$)
                node [align=center, text width=2.5cm, pos=0.5, anchor=west] {#4};
    \end{tikzpicture}
}%

\begin{document}
\mainmatter              

\title{Operation and Imitation under \\ Safety-Aware Shared Control}

\titlerunning{Safety-Aware Shared Control}  

\author{Alexander Broad\inst{1}\inst{2} \and Todd Murphey\inst{1}\and
Brenna Argall\inst{1}\inst{2}}

\authorrunning{Broad, Murphey and Argall} 
%
\tocauthor{Alexander Broad, Todd Murphey, Brenna Argall}
\institute{Northwestern University, Evanston IL 60208, USA\\
\and
Shirley Ryan AbilityLab, Chicago, IL 60611, USA\\
\email{alex.broad@u.northwestern.edu}}

\maketitle              

\begin{abstract}
We describe a shared control methodology that can, without knowledge of the task, be used to improve a human's control of a dynamic system, be used as a training mechanism, and be used in conjunction with Imitation Learning to generate autonomous policies that recreate novel behaviors.  Our algorithm introduces autonomy that assists the human partner by enforcing safety and stability constraints.  The autonomous agent has no a priori knowledge of the desired task and therefore only adds control information when there is concern for the safety of the system.  We evaluate the efficacy of our approach with a human subjects study consisting of 20 participants.  We find that our shared control algorithm significantly improves the rate at which users are able to successfully execute novel behaviors.  Experimental results suggest that the benefits of our safety-aware shared control algorithm also extend to the human partner's understanding of the system and their control skill.  Finally, we demonstrate how a combination of our safety-aware shared control algorithm and Imitation Learning can be used to autonomously recreate the demonstrated behaviors.
\end{abstract}

\section{Introduction}
\label{sec-intro}

Mechanical devices can be used to extend the abilities of a human operator in many domains, from travel to manufacturing to surgery.  In this work, we are interested in developing a shared control methodology that further enhances the ability of a human user to operate dynamic systems in scenarios that would otherwise prove challenging due to the complexity of the control problem (e.g., modern aircraft), the complexity of the environment (e.g., navigation in a densely populated area), or the skill of the user (e.g., due to physical injury).  A particularly motivating domain is assistive and rehabilitation medicine.  Consider, for example, the use of an exoskeleton in rehabilitating the leg muscles of a spinal cord injured subject~\cite{colombo2000treadmill}.  While these devices are designed explicitly to aid a user in recovering from trauma by rebuilding lost muscular control, the complexity of the machine itself often requires that one or more physical therapists assist the subject in operating the device during therapy (e.g., to provide stabilization).  Artificial intelligence can further improve the efficacy of these devices by incorporating autonomy into the control loop to reduce the burden on the human user.   That is, if the autonomous agent accounts for subpar (and potentially dangerous) control input, the human operator and therapist(s) are freed to focus on important therapeutic skills.

In this work, we improve the effectiveness of joint human-machine systems by developing a safety-aware shared control (SaSC) algorithm that assists a human operator in controlling a dynamic machine without a priori knowledge of the human's desired objective.  In general, shared control is a paradigm that can be used to produce human-machine systems that are more capable than either partner on their own~\cite{music2017control}.  However, in practice, shared control systems often require the autonomous agent to know the goal (or a set of discrete, potential goals). While a priori knowledge of a desired set of goals may be a valid assumption in some domains, it can also be a severely limiting assumption in many other scenarios.  Therefore, instead of allocating control based on whether the human operator's input will improve the likelihood of achieving a goal, we aim to allocate control based on whether the user's control commands will lead to dangerous states and actions in the future.  Under this paradigm, the autonomous partner develops a control strategy that is \textit{only concerned with the safety of the system, and is otherwise indifferent to the control of the human operator.}

Our safety-aware shared control algorithm can be used to improve the efficacy of human-machine systems in two main ways:
\begin{description}
\item[G1.] Improve a human operator's control, and understanding, of a dynamic system in complex and potentially unsafe environments.
\item[G2.] Improve the value of Imitation Learning (IL) in the same domains, both by facilitating demonstration and addressing the covariate shift problem~\cite{ross2011reduction}.
\end{description}
Item \textbf{G1} is important because control challenges can stem from a variety of issues including the inherent complexity of the system, the required fidelity in the control signal, or the physical limitations of the human partner.  For this reason, there is often an explicit need for assistance from an autonomous partner in controlling the mechanical device.  Item \textbf{G2} is important because Imitation Learning can be used to further extend the capabilities of human-machine systems, however demonstration may not always be feasible for the human partner given the aforementioned control challenges.

The main contribution of this work is a safety-aware shared control algorithm that improves the efficacy of human-machine collaboration in complex environments, \textit{with the key feature that it is possible for the user's desired objective to remain unknown}.  In this algorithm, an autonomous partner accounts for system- and environment-based safety concerns, thereby freeing the human operator to focus their mental and physical capacities on achieving high-level goals.  Our algorithm (Section~\ref{sec-sasc}) describes a novel interaction paradigm that extends the viability of complex human-machine systems in various scenarios including those in which the human's skill is limited or impaired.  We also provide an analysis of our algorithm (Section~\ref{sec-empirical-evaluation}) with a human subjects study consisting of 20 participants conducted in a series of challenging simulated environments.  Finally, we show how the same algorithm can be used to improve the human operator's control skill and the power of Imitation Learning (Section~\ref{sec-results}).

\section{Background and Related Work}
\label{sec-background-and-related-work}

The majority of related work in the shared control literature focuses on the development and analysis of techniques that statically and/or dynamically allocate control between human and robot partners.  The main objective in these works is to improve system performance while reducing the control burden on the human operator~\cite{music2017control}.  In some applications, the autonomous partner is allocated the majority of the low-level control while the human operator acts in a supervisory role~\cite{aigner2000modeling} while in others, the roles are reversed, and the autonomous partner takes on the supervisory role~\cite{shen2004collaborative}.  There is also related work in which neither partner acts as a supervisor and instead control is shared via a mixed-initiative paradigm.  For example, researchers have developed techniques in which the autonomous partner is explicitly aware of the human operator's intention~\cite{chipalkatty2013less} as well as techniques in which the autonomous partner has an implicit understanding of the human operator's control policy~\cite{broad2017learning}.  While these examples aim to improve task performance, the main motivation for our work is to extend the human operator's control ability in domains where safety is a primary concern.

For this reason, the most closely related shared control paradigms are safety, or context, aware.  In this area, researchers have explored the impact of autonomous obstacle avoidance on teleoperated robots in search and rescue~\cite{shen2004collaborative}. Additionally, safety is a particular concern when the human and robot partner are co-located, such as with autonomous vehicles~\cite{anderson2010optimal}.  Co-located partners are also common in assistive robotics where researchers have developed environment-aware smart walkers~\cite{lacey2000context} to help visually-impaired people avoid dynamic obstacles.  

Related to our goal of generating autonomous policies that recreate the behavior demonstrated during system operation, there is prior work in the field of Imitation Learning (IL).  Most commonly, the demonstration data is provided by a human partner~\cite{argall2009survey}, though it can come from variety of sources including trajectory optimization and simulation~\cite{levine2013guided}.  Example data is commonly assumed to come from an expert (or optimal) controller~\cite{abbeel2004apprenticeship}; however, researchers also have explored techniques that use demonstrations provided by novice (or suboptimal) sources~\cite{ziebart2008maximum}.  In this work we describe how Imitation Learning can be used even when \textit{the human operator is not able to provide demonstration data on their own}.  We further describe how our safety-aware shared control algorithm can be used to address the covariate shift problem~\cite{ross2011reduction}, a common issue in Imitation Learning that stems from the fact that the data used to train the policy may come from a different distribution than the data observed at runtime.

Lastly, there is also related work in the subfield of safety-aware Reinforcement Learning (RL).  In this domain, safe autonomous control policies are learned from exploration instead of demonstration.  Examples include techniques that enforce safety during the exploration process through linear temporal logic~\cite{alshiekh2017safe} and formal methods~\cite{fulton2018safe}. Researchers also have explored model-free RL as a paradigm to integrate the feedback of a human operator through the reward structure during the learning process, and to share control with a human operator at run-time~\cite{reddy2018shared}.  Finally, researchers have considered learning safe policies from demonstration by computing probabilistic performance bounds that allow an autonomous agent to adaptively sample demonstration data to ensure safety in the learned policy~\cite{brown2018efficient}.  

\section{Safety-Aware Shared Control}
\label{sec-sasc}
\begin{figure*}[!t]
	\centering
	\includegraphics[width=0.95\linewidth]{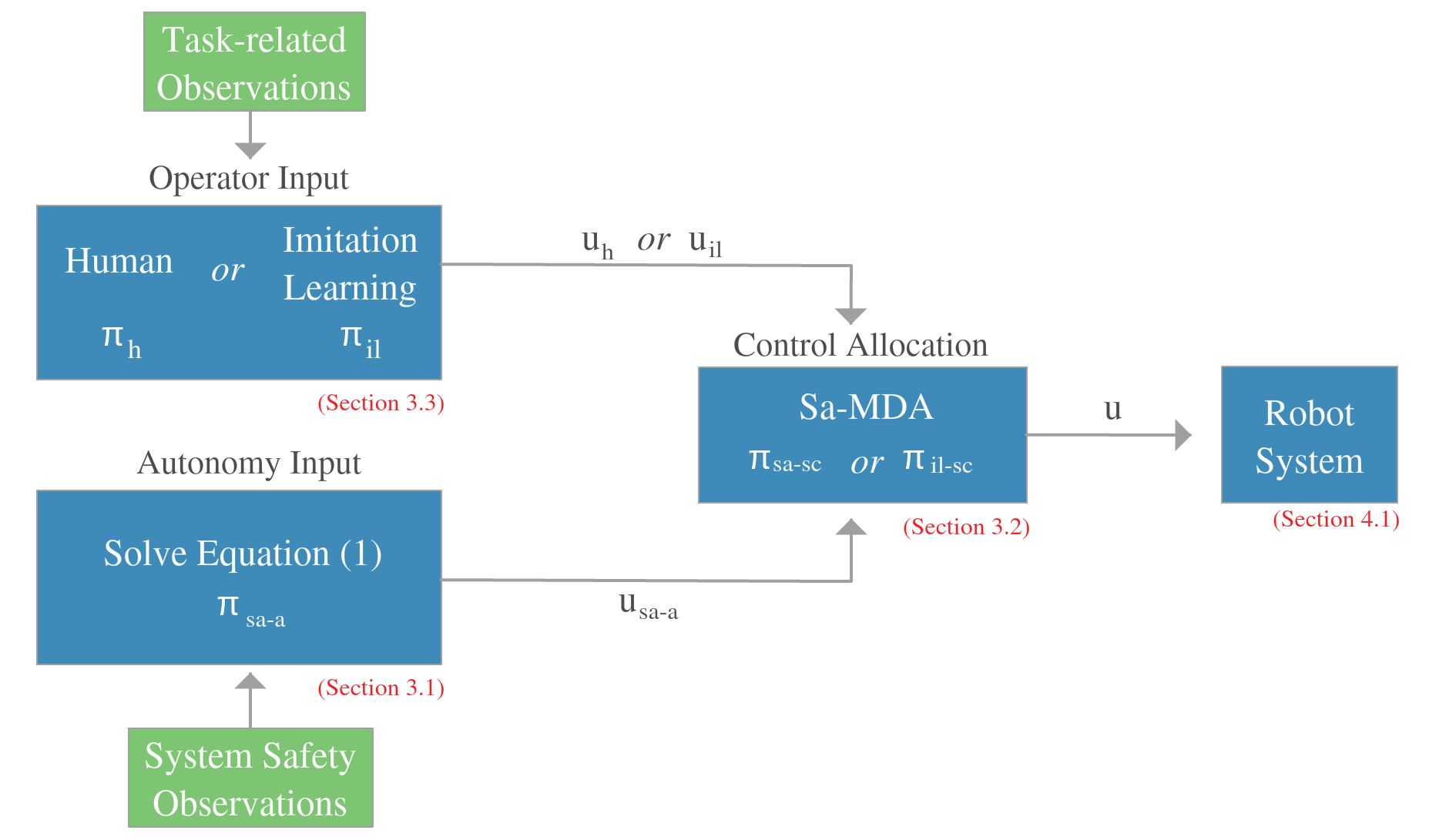}
	\caption{Flow chart of our Safety-aware Shared Control (SaSC) algorithm.  The operator  focuses on task objectives, while the autonomy accounts for safety.}
	\label{fig-sasc}
\end{figure*} 

In this work, we contribute a specific implementation of a class of algorithms that we refer to as Safety-aware Shared Control (SaSC).  SaSC can help users operate dynamic systems in challenging and potentially unsafe environments that would normally require expert-level control.  An analogous engineering solution is fly-by-wire control of modern aircraft~\cite{sutherland1968fly}.  In these systems, the onboard computer accounts for many of the intricacies of the control problem allowing the pilot to focus on high-level tasks.  Our SaSC algorithm takes this idea a step further to account for unsafe actions related to both the system dynamics and the environment.  Safety-aware shared control consists of two components:
\begin{description}
\item[1.] A safe policy generation method for the autonomous agent.
\item[2.] A control allocation algorithm to safely integrate input from both partners.
\end{description}
The high-level algorithm is depicted in Figure~\ref{fig-sasc}. Points \textbf{1} and \textbf{2} are described in more detail in subsections ~\ref{subsec-model-based-optimal-control} and ~\ref{subsec-dynamic-shared-control}.  Relevant policy notation is below:
\begin{itemize}
\item $\pi_h \: \: \: \: \: \: \: \:$ Human operator's policy
\item $\pi_{sa-a} \: \:$ Safety-aware autonomous policy (has no task information)
\item $\pi_{sa-sc} \:$ Safety-aware shared control policy
\item $\pi_{il} \: \: \: \: \: \: \: \:$ Imitation Learning policy
\item $\pi_{il-sc} \: \:$ Imitation Learning policy under safety-aware shared control
\end{itemize}

\subsection{Safety-Aware Autonomous Policy Generation}
\label{subsec-model-based-optimal-control}

To implement our safety-aware shared control algorithm, we must first develop an autonomous control policy that is capable of \textit{safely controlling} the dynamic system in question (policy $\pi_{sa-a}$ in Section~\ref{sec-sasc}).  In this work, we utilize Model-based Optimal Control~\cite{atkeson1997comparison} (MbOC).  MbOC learns a model of the dynamic system directly from data, which is then incorporated into an optimal control algorithm to produce autonomous policies.  Here, we learn a model of the system and control dynamics through an approximation to the Koopman operator~\cite{koopman1931hamiltonian}.  Further details of this modeling technique are presented in Section~\ref{sec-sub-sub-model-learning}.

Given a learned model of the system dynamics, we can then compute a control policy by solving the finite-horizon nonlinear Model Predictive Control (MPC)~\cite{rawlings2017model} problem defined by a cost function of the form
\begin{equation}
J(x(t), u(t)) = \int_{t=0}^{t_f} l(x(t),u(t)) + l_{t_f}(x(t))
\label{eq-cost}
\end{equation}
\noindent where
\begin{equation}
\dot{x}(t) = f(x(t),u(t)),\quad x(0) = x_0
\label{eq-sys-dynamics}
\end{equation}
\noindent and $f$ defines the nonlinear system dynamics, $x(t)$ and $u(t)$ are the state and control trajectories, and $l$ and $l_{t_f}$ are the running and terminal costs, respectively.  

To solve the optimal control problem, we use Sequential Action Control (SAC)~\cite{ansari2016sequential}, an algorithm designed to iteratively find a single action (and time to act) that maximally improves performance.  Data-driven model-based, and model-free, optimal control algorithms have been experimentally validated with numerous dynamic systems~\cite{abraham2017model} including joint human-machine systems~\cite{broad2017learning}~\cite{reddy2018shared}.

To address the main focus of this work, \textit{we specify a control objective that relates only to system safety}.  Specifically, we use quadratic costs to deter unstable states and higher order polynomial penalty functions to keep the system away from dangerous locations.  Notably, the cost function \textit{does not} incorporate any task information.  Therefore, if the autonomous partner's policy is applied directly, the system will work to maintain a safe state but will not move towards any specific goal.  The cost function used in our work is described in Section~\ref{sub-sec-implementation-autonomous-policy-generation}.

\subsection{Safety-Aware Dynamic Control Allocation}
\label{subsec-dynamic-shared-control}

To assist the human partner in safely controlling the dynamic system, we define an outer-loop control allocation algorithm that incorporates input signals from the human and autonomous partners (policy $\pi_{sa-sc}$ in Section~\ref{sec-sasc}).  There is, of course, a balance to strike between the control authority given to the each partner.  If the outer-loop controller is too permissive and accepts a significant portion of the human operator's input, it may do a poor job enforcing the necessary safety requirements.  However, if the outer-loop controller is too stringent, it can negatively impact the ability of the human operator to produce their desired motion. In this work, we balance the control authority between the human and autonomous partners to increase the authority of the human operator when the system is deemed to be in a safe state, and increase the authority of the autonomy when the system is deemed to be in a dangerous state.  Here, the autonomy \textit{adds information into the system only when it is necessary to ensure safety}.  

Specifically, we allocate control using a variant of Maxwell's Demon Algorithm (MDA)~\cite{tzorakoleftherakis2015controllers}.  MDA uses information from an optimal control algorithm as a guide by which to evaluate the input from another source.  In this work, we contribute a safety-aware variant that we call Safety-Aware MDA (Sa-MDA).  Sa-MDA is described in full in Algorithm~\ref{alg-mda}.  Here, $\textbf{unsafe}$(system,environment) describes whether the system is in an unstable or dangerous configuration with respect to the environment (e.g., through barrier functions, see Section~\ref{sub-sec-implementation-autonomous-policy-generation}), $\boldsymbol{u_h}$ is the input from the human partner, $\boldsymbol{u_a}$ is the input produced by the autonomy, $\langle \cdot \rangle$ is the inner product, and $\boldsymbol{u}$ is the applied control.  When a learned policy is used to mimic a demonstration $\boldsymbol{u_h}$ is replaced with $\boldsymbol{u_{il}}$ (see Section~\ref{sec-sub-sasc-il}).

\begin{algorithm}[!t]
 	\footnotesize
	\begin{algorithmic}[1]
		\If {\textbf{unsafe} (system, environment)}  
			\State $\boldsymbol{u} = \boldsymbol{u_{sa-a}}$;  
		\Else
			\If {$\langle \boldsymbol{u_h}, \boldsymbol{u_{sa-a}} \rangle \geq 0$} \tikzmark{right} \tikzmark{top} \label{marker} \Comment{When used with an IL policy}
				\State $\boldsymbol{u} = \boldsymbol{u_h}$; \Comment{$u_h$ is replaced by $u_{il}$}
			\Else
				\State $\boldsymbol{u} = \boldsymbol{0}$;
			\EndIf \tikzmark{bottom}
		\EndIf
	\end{algorithmic} 
	\AddNote{top}{bottom}{right}{Maxwell's \\ Demon\\ Algorithm}	
	\caption{Safety-Aware Maxwell's Demon Algorithm}
	\label{alg-mda}
\end{algorithm}

\subsection{Safety-Aware Shared Control Imitation Learning}
\label{sec-sub-sasc-il}

We now describe how we use the data collected under shared control to produce autonomous policies through Imitation Learning~\cite{argall2009survey}.  The goal here is to learn a policy $\pi_{il}$ that mimics the behavior demonstrated by the human operator.  To achieve this goal, we treat the data collected under shared control $\pi_{sa-sc}$ as a supervisor in the policy learning process.  Notably, this data, and the associated learned policy, \textit{now contain task-relevant information}, as provided by the human operator during demonstration.  Our goal, then, is to learn an autonomous policy that minimizes the following objective
\begin{equation}
J(\pi_{il}, \pi_h) = \min_\theta \sum_{s \in \xi \in \mathbb{D}}|| \pi_{il}(s) - \pi_h(s) ||_2^2 
\label{eq-lfd}
\end{equation}

\noindent where $J$ is the cost to be minimized and $s$ is the state.  By minimizing $J$ we learn a policy that closely matches the policy demonstrated by the human partner.  $\pi(s) : s \rightarrow u$ defines a control policy which is parameterized by $\theta$, $\pi_h$ represents the (supervisor) human's policy and $\pi_{il}$ represents the Imitation Learning policy.  

The autonomous policy is learned from a set $\mathbb{D}$ of trajectory data $\xi$ recorded during the demonstration phase.  To generate $\pi_{il}$, we use behavior cloning~\cite{laskey2017dart}, a standard offline imitation learning algorithm.  Details of our specific implementation are provided in Section~\ref{sec-implementation-details}.  As described in Section~\ref{sec-background-and-related-work}, behavior cloning can fail to reproduce the desired behavior due to the covariate shift problem~\cite{ross2011reduction}~\cite{laskey2017dart}.  We address this issue with an autonomous policy $\pi_{il-sc}$ that \textit{combines} the learned policy $\pi_{il}$ with the safety-aware autonomous policy $\pi_{sa-a}$ (Algorithm~\ref{alg-mda}).   By incorporating the same shared control algorithm used during data collection, we encourage the system to operate in a similar distribution of the state space to what was observed during demonstration.  One can view this solution as a shared control paradigm in which the control is shared between \textit{two autonomous agents}: the autonomy mimicking the human control and the autonomy enforcing safety constraints. 

\section{Empirical Evaluation}
\label{sec-empirical-evaluation}

To evaluate the efficacy of our algorithm, we perform a human subjects study on a simulated system exhibiting nonlinear dynamics in complex environments.  In this evaluation, we compare the efficacy of each policy presented in Section~\ref{sec-sasc} except $\pi_{sa-a}$, which is never executed and is only used to keep $\pi_h$ and $\pi_{il}$ safe.

\subsection{Experimental System}
\label{sec-experimental-system}

\begin{figure*}[!b]
	\centering
	\includegraphics[width=0.305\linewidth]{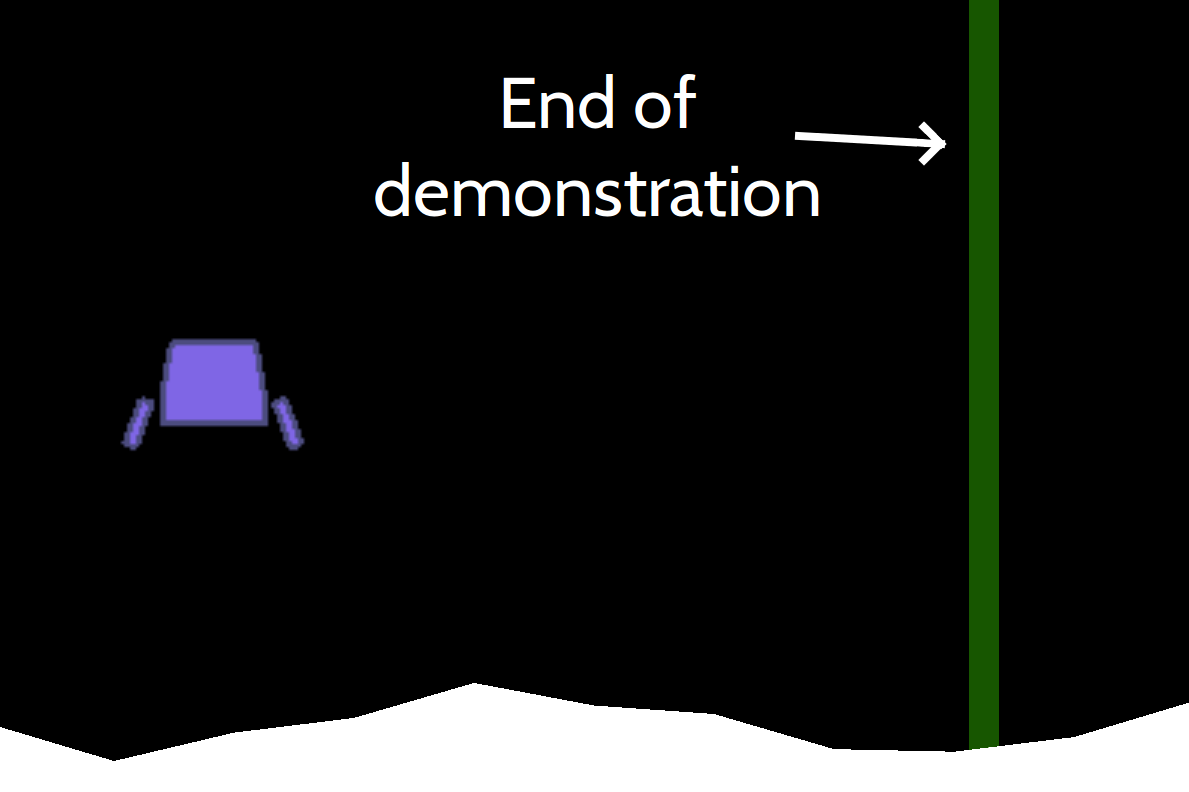} \hspace{0.3cm}
	\includegraphics[width=0.305\linewidth]{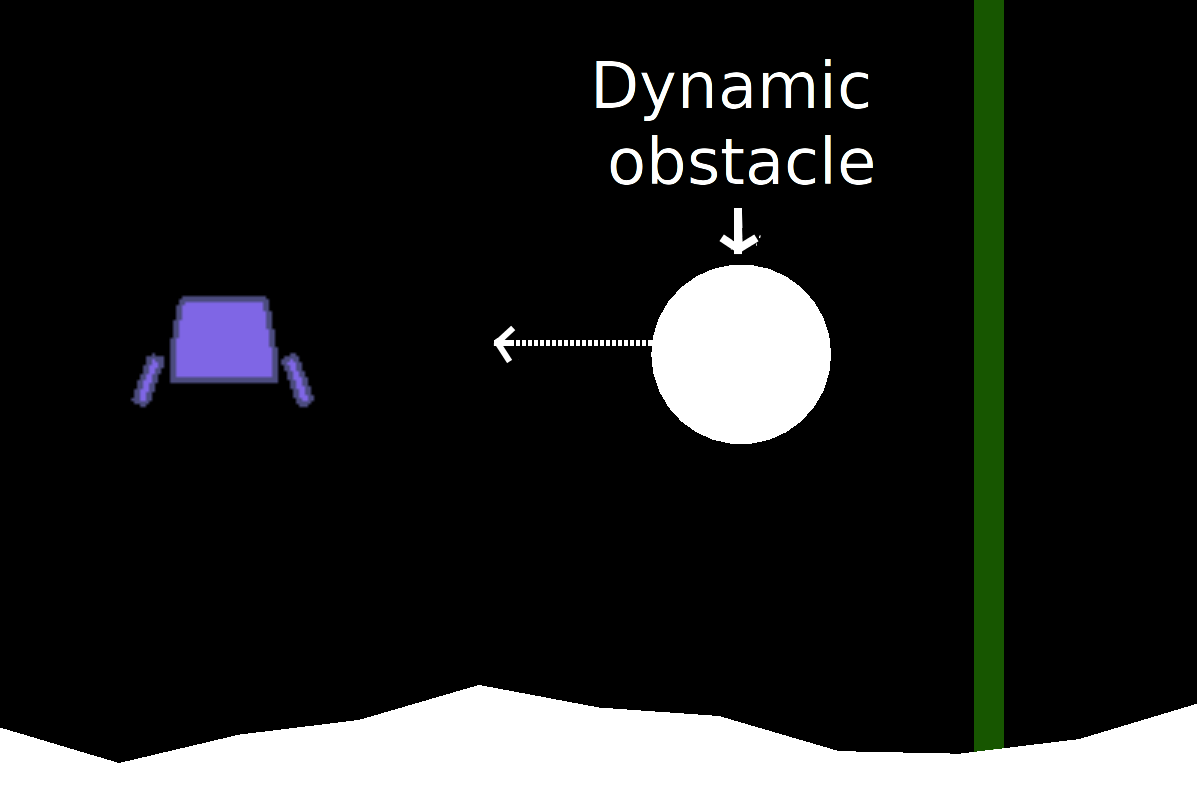} \hspace{0.3cm}
	\includegraphics[width=0.305\linewidth]{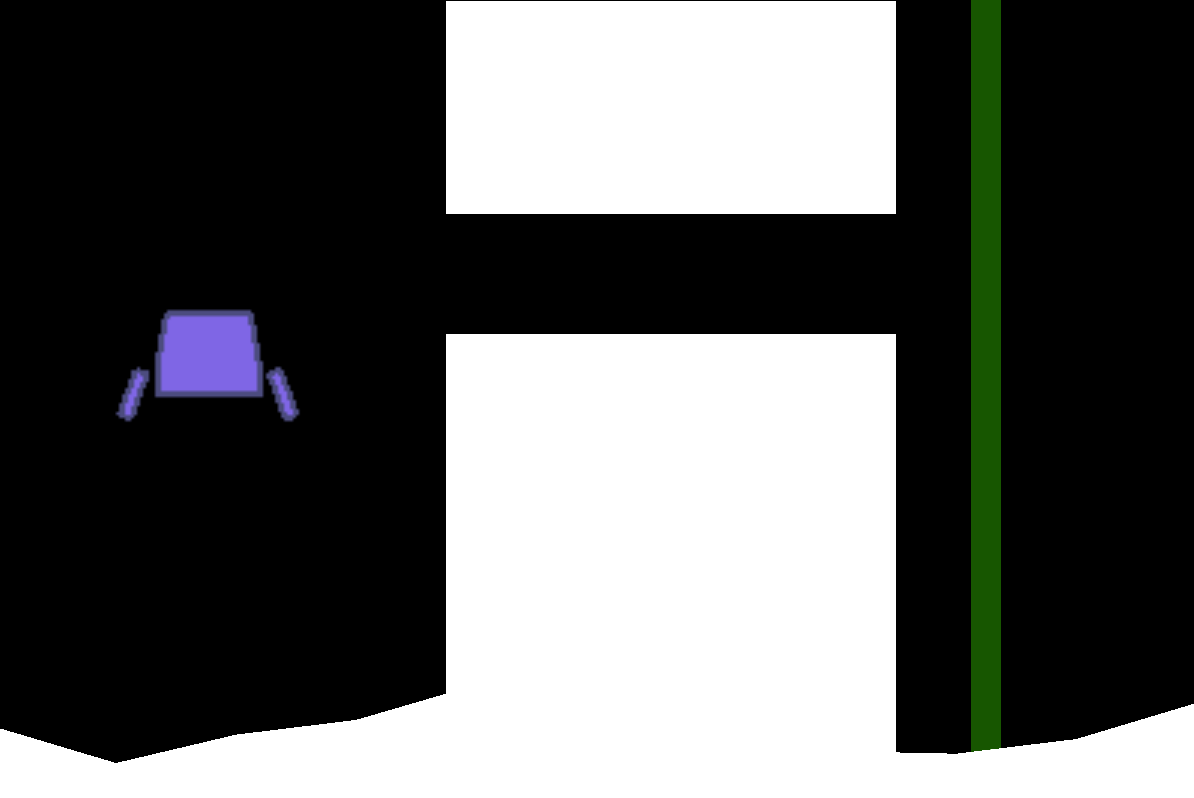}
	\caption{Visualization of lunar lander (enlarged) and experimental environments.  A trial is complete when the lander moves across the green line boundary.}
	\label{fig-envs}
\end{figure*} 

The experimental system consists of a simulated ``lunar lander'' (Figure~\ref{fig-envs}), chosen to demonstrate the impact that shared control can have on the safety of a joint human-machine system when the control problem and environment are complex.  The lunar lander exhibits nonlinear dynamics and can easily become unstable as it rotates away from its point of equilibrium.  Additionally, two of the experimental environments contain obstacles that must be avoided to stay safe (see Figure~\ref{fig-envs}).  The system and environment are implemented in the Box2D physics engine based on the environment defined in OpenAI's Gym~\cite{brockman2016gym}.  

The lunar lander is defined by a six dimensional vector which includes the 2D position and heading ($x_{1-3}$) and their rates of change ($x_{4-6}$).  The control input is a continuous two dimensional vector which represents the throttle of the main ($u_1$) and rotational ($u_2$) thrusters.  The \textit{first} environment includes only the lunar lander and the ground surface (Fig.~\ref{fig-envs}, left).  This environment illuminates the challenges associated with maintaining the stability of a complex dynamic system, while simultaneously executing unspecified behaviors.  The \textit{second} environment incorporates dynamic obstacles that obstruct the motion of the system (Fig.~\ref{fig-envs}, middle).  In this environment, a series of circular obstacles move across the screen at the same height as the lander (one at a time).  The \textit{third} environment includes two static obstacles that force the operator to navigate through a narrow passageway, increasing the required control fidelity (Fig.~\ref{fig-envs}, right).  

\subsection{Implementation Details}
\label{sec-implementation-details}

\subsubsection{Model Learning}
\label{sec-sub-sub-model-learning}

We learn a model of the system and control dynamics using an approximation to the Koopman operator~\cite{williams2015data}, which has been validated on numerous systems~\cite{abraham2017model}, including human-machine systems~\cite{broad2017learning}.  The Koopman is a linear operator that can model all relevant features of nonlinear dynamical systems by operating on a nominally infinite dimensional representation of the state~\cite{koopman1931hamiltonian}.  To approximate the true Koopman operator one must define a basis.  In this work, we define $\phi = [1, x_1, x_2, x_3, x_4, x_5, x_6, u_1, u_2, u_1\cdot x_1, u_1\cdot x_2, u_1\cdot x_3, u_1\cdot x_4, u_1\cdot x_5, u_1\cdot x_6, u_2\cdot x_1, u_2\cdot x_2, u_2\cdot x_3, u_2\cdot x_4, u_2\cdot x_5, u_2\cdot x_6, u_1\cdot cos(x_3), u_1\cdot sin(x_3), u_2\cdot cos(x_3), u_2\cdot sin(x_3)]$, where $x_{1-6}$ represent the system state variables and $u_{1-2}$ represent the control input variables.  The specific basis elements used in this work are chosen empirically and represent a reduced set of features that describe the full state and control space, as well as the interaction between the user's input and the state.  Data-driven methods (e.g. sparsity-promoting DMD~\cite{jovanovic2014sparsity}) can be used to automatically choose a proper set of basis functions.

\subsubsection{Safety-Aware Autonomous Policy Generation}
\label{sub-sec-implementation-autonomous-policy-generation}

To compute an autonomous policy that is solely concerned with the safety of the system, we define a cost function (Equation~\eqref{eq-cost}) that considers two notions of safety: stability around points of equilibrium and collision avoidance,
\begin{align*}
l&(x)=\\
&\overbrace{\textstyle Diag[0, 0, 15\cdot x_3, 1\cdot x_4, 1\cdot x_5, 10\cdot x_6]^2}^{\text{stabilization}} + 
\overbrace{\textstyle Diag[(x_1-o_1), (x_2-o_2), 0, 0, 0, 0]^8}^{\text{obstacle avoidance}}
\end{align*}
\noindent where $(o_1, o_2)$ is the position of the nearest obstacle in 2D space.  This cost function (i) penalizes states that are far from points of equilibrium using a quadratic cost and (ii) prevents the system from entering dangerous portions of the state space using polynomial barrier functions~\cite{boyd2004convex}.  The stabilization term ensures that the lunar lander does not rotate too far away from upright---if this happens, the lander's main thruster can no longer be used to counteract gravity, a situation that commonly leads to catastrophic failure in rockets.  We therefore penalize both the position and velocity of the heading term.  We additionally penalize the $x$ and $y$ velocity terms as momentum can significantly reduce the time a controller has to avoid collision.  Finally, the obstacle avoidance term simply acts to repel the system from the nearest obstacle.  Importantly, if this policy is applied on its own (without any input from the human partner), the lander will simply attempt to hover in a safe region, and will not advance towards any goal state. 

Notably, the safety-aware autonomous policy only \textit{adds} information into the control loop when the system is deemed to be \textbf{unsafe} (see Algorithm~\ref{alg-mda}).  Otherwise the user's commands are, at most, simply blocked by the autonomy.  This can be thought of as an accept-reject-replace shared control paradigm~\cite{kalinowska2018online}.  We define the $\textbf{unsafe}$ function based on an empirically chosen distance to the nearest obstacle.  Therefore, if the system gets too close to an obstacle, the autonomy's signal is sent to the system, otherwise the human's input is accepted or rejected, according to the MDA filter.  Here we note that the structure and weights in the defined cost function, as well as the pre-defined distance metric, are specific to the experimental system; however, there are generalizable principles that can be used to develop similar cost function for other systems.  For example, system stability can generally be improved by defining costs that help reduce dynamic features to kinematic features, while obstacle avoidance terms can be defined using additional information from the learned system model.

\subsubsection{Imitation Learning}
\label{sec-sub-il-alg}

A neural network is used to learn a control policy that mimics successful trials demonstrated by the human partner.  The input is the current state ($x_{1-6}$) and the output is the control signal ($u_{1-2}$) sent to the system.  The control signal is discretized ($-1.0$ to $1.0$ in increments of $0.5$) and the problem is therefore cast as a classification instead of regression.  There are three hidden layers in the neural network---the first has 32 nodes, and the following two layers have 64 nodes.  Each hidden layer uses ReLu as an activation function.  The final layer uses a softmax activation.  We use categorical cross entropy to compute the loss and RMSProp as the optimizer.

\subsection{Study Protocol}

The human subjects study consisted of 20 participants (16 female, 4 male).  All subjects gave their informed consent and the experiment was approved by Northwestern University's Institutional Review Board.  Each participant provided demonstrations of novel behaviors in all three of the environments, under both a \textit{user-only control} paradigm and our \textit{safety-aware shared control} paradigm.  There was no goal location specified to the participants; instead a trial was considered complete when the human operator navigated the lunar lander across a barrier defined by the green line in the environment (see Figure~\ref{fig-envs}).  The specific trajectory taken by the lander during a demonstration was up to the participant.  The operator used a PS3 controller to interact with the system.  The joystick controlled by the participant's dominant hand fired the main thruster, and the opposing joystick fired the side thrusters.  

Subjects were asked to provide 10 demonstrations per environment (3 total) and per control paradigm (2 total), resulting in a total of 60 demonstrations per participant.  The environments were presented in a randomized and counterbalanced order.  Participants were assigned to one of two groups, where Group A (10 subjects) provided demonstration data in each environment under \textit{user-only control} first, and Group B (10 subjects) provided demonstration data under \textit{shared control} first.  Group assignment was random and balanced across subjects.

\section{Experimental Results}
\label{sec-results}

We find that our SaSC algorithm significantly improves the user's skill with respect to a no-assistance baseline (Figure~\ref{fig-res-main}).  Additionally, we find that our SaSC algorithm can be used as a training mechanism to improve a subject's understanding and control of the dynamic system (Table~\ref{fig-res-tertiary}).  Finally, we show how the same shared control technique can be used to extend the Imitation Learning paradigm (Figure~\ref{fig-lfd-env-2} and~\ref{fig-lfd-env-3}).  These findings are discussed in detail in the following subsections.

\subsection{Safety-Aware Shared Control Enables Successful Demonstration}
\label{sec-sub-success-rate}

\begin{figure*}[!b]
	\centering
	\includegraphics[width=\linewidth]{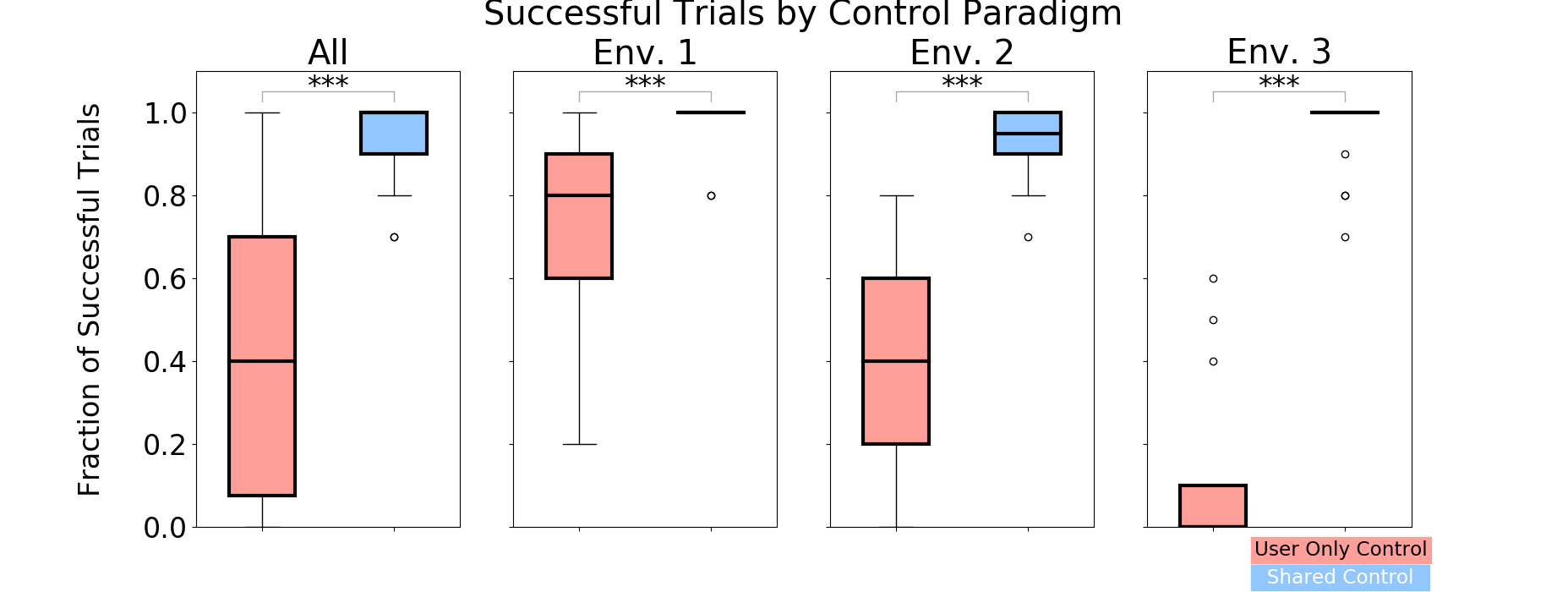} 
	\caption{Average fraction of successful trials under each control paradigm in each  environment. The plots represent data collected in all environments (left), and broken down by each individual environment (right).  In all cases, participants under the shared control paradigm provide safe demonstrations significantly more often than under the user-only control paradigm (*** : $p < 0.005$).}
	\label{fig-res-main}
\end{figure*} 

To address item \textbf{G1}, the primary metric we evaluate is a binary indicator of control competency: the occurrence of safe, successful demonstrations, indicated by navigating the lunar lander beyond the green border.  We first analyze data collected from all three experimental environments together.  We then segment the data based on the specific environment in which it was collected and re-perform our analysis (Figure~\ref{fig-res-main}).  We use the non-parametric Wilcoxon signed-rank test to statistically analyze the data.

The results of the statistical tests revealed that our described shared control paradigm significantly improved the human partner's control skill ($p < 0.005$).  In particular, we find that participants provided safe demonstrations of the desired behavior in $96.0\%$ of the trials produced under the shared control paradigm versus $38.5\%$ of the trials produced under the user-only control paradigm.  Additionally, the statistically significant result holds when we compare the control paradigms in each experimental environment separately ($p < 0.005$ in all cases).  

Recall that the shared control paradigm does not provide any \textit{task}-related assistance, but rather only \textit{safety}-related assistance.  We therefore interpret the increase in task success as evidence that our SaSC algorithm helps subjects exhibit greater control skill, and an associated increased ability to provide demonstrations of novel behaviors.  Additionally, we note that in some experiments conducted under user-only control, subjects were \textit{not able to provide any successful demonstrations} in a given environment.  This suggests that safety-aware shared control may be a \textit{requirement} for functional usage of dynamic systems in some of the more challenging domains that motivate this work.  This finding is also important when considering our ability to train autonomous policies that mimic behaviors demonstrated by the human operator (see Section~\ref{sec-sub-sasc-improves-il}).

\subsection{Impact of Safety-Aware Shared Control on Trajectory Features}
\label{sec-sub-baseline}

\begin{table}[!b]
\centering
\begin{tabular}{|c|c|c|c|c|}
\hline
\textbf{Metric} & \textbf{Control} & \textbf{Env 1.} & \textbf{Env 2.} & \textbf{Env 3} \\
\hline
\multirow{ 2}{*}{\textbf{Path Length (m)}} & User & 30.0 \textpm 2.4 & 33.8 \textpm 4.8 & 28.3 \textpm 0.8 \\
 & Shared & 27.7 \textpm 1.1 & 34.4 \textpm 2.3 & 30.5 \textpm 2.2\\
\hline
\multirow{ 2}{*}{\textbf{Trial Time (s)}} & User & 15.5 \textpm 3.5 & 18.6 \textpm 5.6 & 22.4 \textpm 5.9 \\
 & Shared & 27.0 \textpm 7.8 & 30.0 \textpm 7.4 & 30.4 \textpm 6.8\\
\hline
\multirow{ 2}{*}{\textbf{Final Speed (m/s)}} & User & 23.6 \textpm 6.8 & 23.8 \textpm 6.2 & 14.5 \textpm 5.0\\
 & Shared & 7.9 \textpm 2.8 & 9.6 \textpm 4.5 & 7.9 \textpm 1.9\\
\hline
\multirow{ 2}{*}{\textbf{Final Heading (deg)}} & User & 45.5 \textpm 73.5 & 79.5 \textpm 79.8 & 39.6 \textpm 46.0 \\
 & Shared & 2.7 \textpm 10.4 & 6.4 \textpm 22.0 & 3.5 \textpm 5.9\\
\hline
\end{tabular}
\vspace{0.1cm}
\caption{Mean and standard deviation of trajectory metrics computed from successful demonstrations.  Path length is not impacted significantly by SaSC, but shared control does result in trajectories that take longer to execute, have slower final speeds, and are more upright (stable) in their final configurations.}
\label{fig-res-secondary}
\end{table}

As discussed in Section~\ref{subsec-dynamic-shared-control}, safety-aware shared control impacts features of the trajectories produced by the human operator.  For example, by rejecting a majority of the user's inputs the SaSC algorithm can ensure system safety, but it will not allow the user to execute desired behaviors.  To evaluate how our SaSC algorithm impacts the user's abilities to demonstrate a novel behavior, we compare a number of quantitative metrics that go beyond safety and relate specifically to features of the trajectories (Table~\ref{fig-res-secondary}).  Here, we analyze only the \textit{successful demonstrations} provided under each control paradigm.

Our analysis shows that the safety-aware shared control paradigm impacts not only the \textit{ability} of a user to provide demonstrations, but also \textit{how} they provide demonstrations.  In all environments, we find that participants produced trajectories of nearly equal length under both control paradigms.  However, under the user-only control paradigm participants produced successful demonstrations in \textit{less time}, with a \textit{greater final speed} and in a state that is \textit{rotated further away from the point of equilibrium} than under the shared control paradigm.  Moreover, in Environments 1 and 2 (the less constrained environments), participants were able to produce demonstrations that were safe over the course of the demonstrated trajectory, \textit{but unstable in the final configuration}.  While these were counted as successful demonstrations, they require less control skill than trajectories that end in a stable configuration.  This suggests that SaSC improves control skill in ways not fully captured by the binary success metric.

\subsection{Safety-Aware Shared Control as a Training Mechanism}
\label{sub-sec-training}

As a final piece of analysis addressing item \textbf{G1}, we examine whether experience under a safety-aware shared control algorithm improves human skill learning.  To evaluate this idea, we compare the user-only control trials of Group A (\textit{user-only condition first}) with those of Group B (\textit{user-only control condition second}).  We segment the data based on the specific environment in which it was collected and use the non-parametric Mann-Whitney U test to statistically analyze the results.  We display data and the results of the described statistical tests in Table~\ref{fig-res-tertiary}.

\begin{table}[!b]
\centering
\begin{tabular}{|l|c|c|c|c|}
\hline
 & \textbf{User-Only First} & \textbf{User-Only Second} & \textbf{Difference} & \textbf{Stat. Significance} \\
\hline
\textbf{Env 1.} & 67 $\%$ & 77 $\%$ & +10 $\%$ & $p > 0.05$\\
\textbf{Env 2.} & 35 $\%$ & 44 $\%$ & +9 $\%$ & $p > 0.05$\\
\textbf{Env 3.} & 0 $\%$ & 18 $\%$ & +18 $\%$ & $p < 0.005$\\
\textbf{All} & 34 $\%$ & 46 $\%$ & +12 $\%$ & $p = 0.06$ \\
\hline
\end{tabular}
\caption{Average success rate under \textit{user-only} control over time.}
\label{fig-res-tertiary}
\end{table}

The important take-away from these results is that shared control allows users \textit{to operate human-machine systems safely during their own skill learning}, and that this practice then \textit{translates to skill retention when the assistance is removed}.  In the most challenging environment, Environment 3, we see the largest raw difference ($+18\%$) in success rate between the two cohorts and a statistically significant result ($p < 0.005$).  Notably, \textit{zero} participants provided \textit{any} successful demonstrations in this environment when the data was provided under the user-only control paradigm first.  We also see a relatively large, but not statistically significant, difference in raw percentage points in Environments 1 and 2 ($+10\%$ and $+9\%$) which might become statistically significant with more data. 

Under this paradigm, we can allow users to learn naturally while simultaneously ensuring the safety of both partners.  For systems where failure during learning is not acceptable (e.g., exoskeleton balancing), safety-aware shared control becomes a requirement in enabling human skill acquisition: without some sort of safety assistance, operators are simply not able to control the system.

\subsection{Safety-Aware Shared Control Improves Imitation Learning}
\label{sec-sub-sasc-improves-il}

\begin{figure}[!b]
	\centering
	\includegraphics[width=0.8\textwidth]{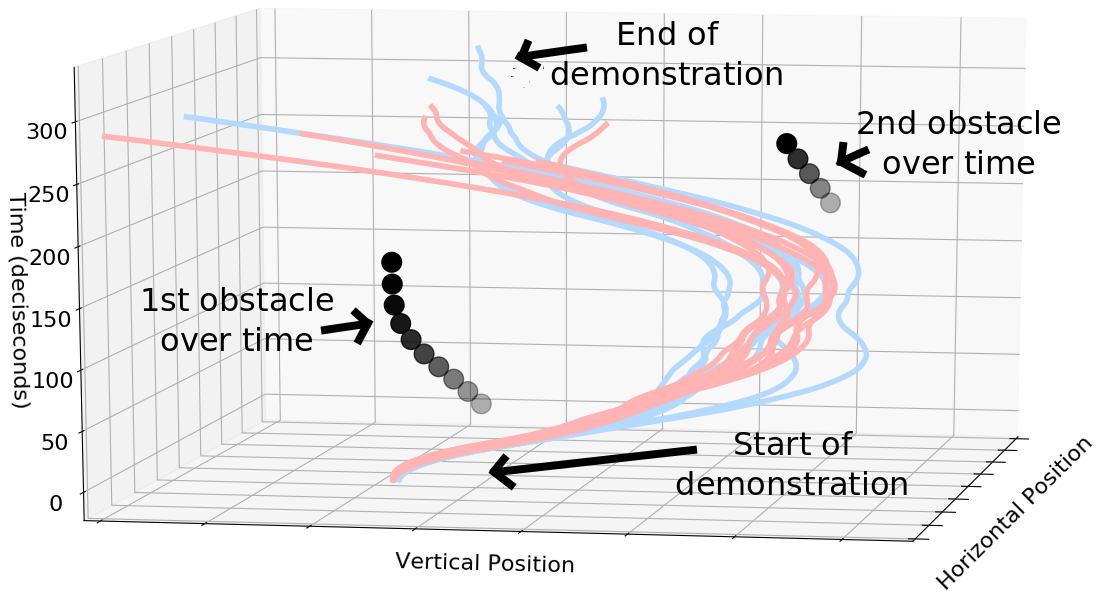}
	\caption{Environment 2. A visualization of data provided by the human partner under safety-aware shared control ($\pi_{sa-sc}$ : blue) and trajectories produced autonomously ($\pi_{il-sc}$ : pink) using the learned control policy.  The vertical and horizontal position of the dynamic obstacles (black) over time are also displayed.}
	\label{fig-lfd-env-2}
\end{figure}

\begin{figure}[!t]
	\centering
    \includegraphics[width=0.8\textwidth]{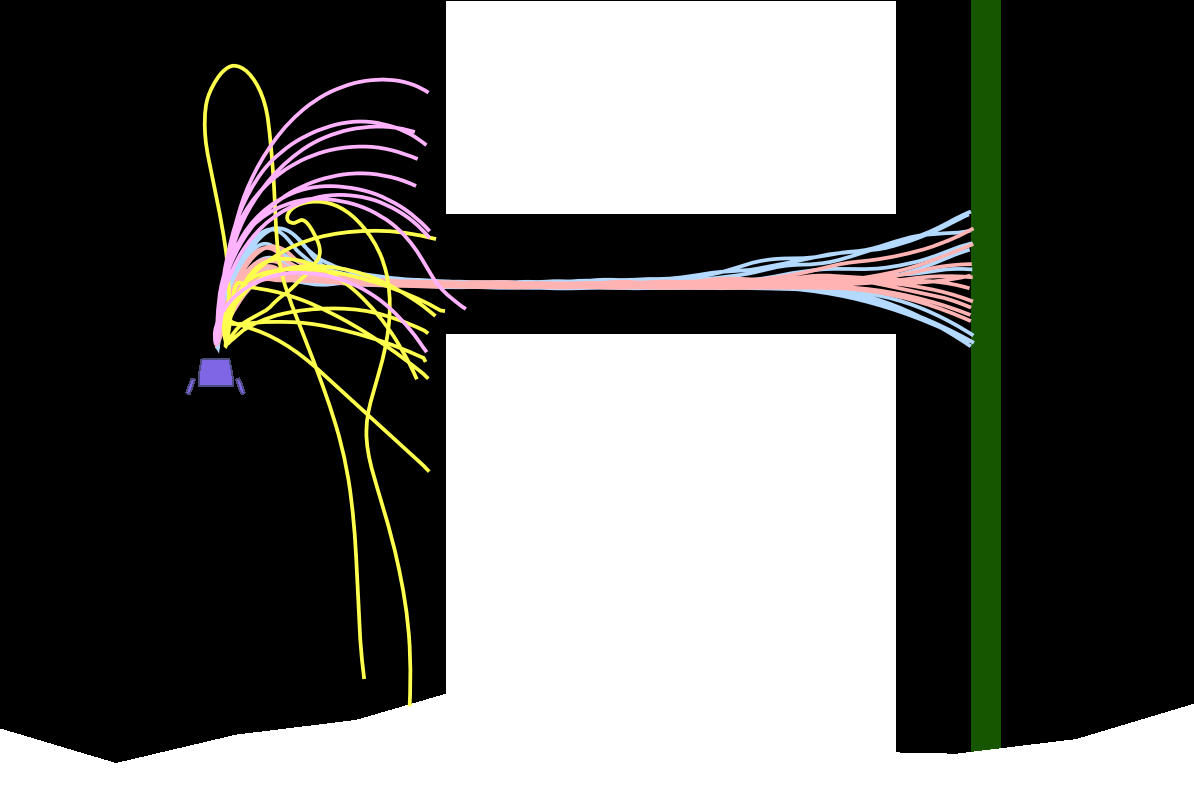}
    \caption{Environment 3.  Visualization of (1) demonstrations provided under the user-only control paradigm ($\pi_{h}$ : yellow), (2) demonstrations provided under our safety-aware shared control paradigm ($\pi_{sa-sc}$ : blue), (3) trajectories produced autonomously using solely the Imitation Learning policy ($\pi_{il}$ : purple, learned from $\pi_{sa-sc}$ demonstrations), and (4) trajectories produced autonomously using the safety aware shared control imitation learning policy ($\pi_{il-sc}$ : pink).}
    \label{fig-lfd-env-3}
\end{figure}

To address item \textbf{G2}, we examine whether learned policies are capable of reproducing the behavior demonstrated by the human operator.  Our evaluation is based on a comparison of trajectories generated by Imitation Learning with ($\pi_{il-sc}$) and without ($\pi_{il}$) safety-aware shared control at runtime.  We provide visualizations of 10 successful reproductions of the demonstrated behaviors in Environments 2 and 3 in Figures~\ref{fig-lfd-env-2} and~\ref{fig-lfd-env-3}, respectively.  

Notably, \textit{all} trajectories produced by $\pi_{il-sc}$ safely avoid both the static and dynamic obstacles.  In Figure~\ref{fig-lfd-env-2} we see that the learned policy $\pi_{il-sc}$ is able to mimic the behavior demonstrated by the human operator.  In Figure~\ref{fig-lfd-env-3}, we include visualizations of the trajectories provided under user-only control ($\pi_{h}$) and Imitation Learning without shared control ($\pi_{il}$) in Environment 3.  Here, we also see that the user was unable to provide any successful demonstrations without safety assistance.  Similarly, the learned control policy ($\pi_{il}$) was unable to avoid obstacles in the environment without the safety assistance.  

These two final points elucidate the \textit{need for our safety-aware shared control system in both the demonstration \textbf{and} imitation phases.}  Without assistance from the SaSC algorithm, not only is the human operator unable to demonstrate desired behaviors, but the learned neural network policy fails to generalize.  

\section{Discussion and Conclusion}
\label{sec-discussion}

In this work, we contribute a shared control paradigm that allows users to provide demonstrations of desired actions in scenarios that would otherwise prove too difficult due to the complexity of the control problem, the complexity of the environment, or the skill of the user.  We solve this problem through the application of shared control, allowing a human operator to provide demonstrations while an autonomous partner ensures the safety of the system.  We validate our approach with a human subjects study comprised of 20 participants.  

The results of our human subjects study show that our safety-aware shared control paradigm is able to help human partners provide demonstrations of novel behaviors in situations in which they would otherwise not be able (Figure~\ref{fig-res-main}).  Additionally, we find that our SaSC algorithm can be used as a training mechanism to improve a human operator's control skill by ensuring safety during training (Table~\ref{fig-res-tertiary}).  Furthermore, we find that a combination of Imitation Learning with our safety-aware shared control paradigm produces autonomous policies that are capable of safely reproducing the demonstrated behaviors.  In this work, the autonomous policies are learned from data provided under shared control and for this reason, one must also consider how the autonomy affects the demonstration data.  We find that the shared control paradigm slows the average speed of the system, but generally increases the stability (Table~\ref{fig-res-secondary}).  Of course, it is also possible to learn autonomous policies from the data provide under user-only control.  However, when system safety is a requirement (e.g. co-located human-machine systems), shared control can be thought of as fundamental for allowing users to provide demonstrations of new behaviors.  Allowing a system to fail during demonstration is often an unrealistic assumption with real-world systems.

In future work, we plan to explore additional uses of data collected under a shared control paradigm in learning autonomous policies that do not rely on continued safety assistance (e.g. as seeds in Guided Policy Search~\cite{levine2013guided}).  We also plan to explore a bootstrapped notion of shared control, in which the outer-loop autonomous controller originally considers only the safety of the joint system and then dynamically updates to consider both the safety of the system system and task-level metrics that describe the desired behavior of the human operator.

\paragraph{Acknowledgements.}  This material is based upon work supported by the National Science Foundation under Grants CNS 1329891 \& 1837515. Any opinions, findings and conclusions or recommendations expressed in this material are those of the authors and do not necessarily reflect the views of the aforementioned institutions. 

\bibliographystyle{IEEEtran}
\bibliography{references}

\begin{thebibliography}{10}
\providecommand{\url}[1]{#1}
\csname url@samestyle\endcsname
\providecommand{\newblock}{\relax}
\providecommand{\bibinfo}[2]{#2}
\providecommand{\BIBentrySTDinterwordspacing}{\spaceskip=0pt\relax}
\providecommand{\BIBentryALTinterwordstretchfactor}{4}
\providecommand{\BIBentryALTinterwordspacing}{\spaceskip=\fontdimen2\font plus
\BIBentryALTinterwordstretchfactor\fontdimen3\font minus
  \fontdimen4\font\relax}
\providecommand{\BIBforeignlanguage}[2]{{%
\expandafter\ifx\csname l@#1\endcsname\relax
\typeout{** WARNING: IEEEtran.bst: No hyphenation pattern has been}%
\typeout{** loaded for the language `#1'. Using the pattern for}%
\typeout{** the default language instead.}%
\else
\language=\csname l@#1\endcsname
\fi
#2}}
\providecommand{\BIBdecl}{\relax}
\BIBdecl

\bibitem{colombo2000treadmill}
G.~Colombo, M.~Joerg, R.~Schreier, and V.~Dietz,
  ``\href{https://www.ncbi.nlm.nih.gov/pubmed/11321005}{Treadmill Training of
  Paraplegic Patients using a Robotic Orthosis},'' \emph{Journal of
  Rehabilitation Research and Development}, vol.~37, no.~6, pp. 693--700, 2000.

\bibitem{music2017control}
S.~Musi{\'c} and S.~Hirche,
  ``\href{https://www.sciencedirect.com/science/article/pii/S1367578817301153}{Control
  Sharing in Human-Robot Team Interaction},'' \emph{Annual Reviews in Control},
  2017.

\bibitem{ross2011reduction}
S.~Ross, G.~J. Gordon, and D.~Bagnell,
  ``\href{http://proceedings.mlr.press/v15/ross11a/ross11a.pdf}{A Reduction of
  Imitation Learning and Structured Prediction to No-Regret Online Learning},''
  in \emph{International Conference on Artificial Intelligence and Statistics},
  2011, pp. 627--635.

\bibitem{aigner2000modeling}
P.~Aigner and B.~J. McCarragher,
  ``\href{http://ieeexplore.ieee.org/document/844360/}{Modeling and
  Constraining Human Interactions in Shared Control Utilizing a Discrete Event
  Framework},'' \emph{Transactions on Systems, Man, and Cybernetics}, vol.~30,
  no.~3, pp. 369--379, 2000.

\bibitem{shen2004collaborative}
J.~Shen, J.~Ibanez-Guzman, T.~C. Ng, and B.~S. Chew,
  ``\href{http://ieeexplore.ieee.org/abstract/document/1438902/}{A
  Collaborative-Shared Control System with Safe Obstacle Avoidance
  Capability},'' in \emph{IEEE International Conference on Robotics, Automation
  and Mechatronics}, 2004, pp. 119--123.

\bibitem{chipalkatty2013less}
R.~Chipalkatty, G.~Droge, and M.~B. Egerstedt,
  ``\href{http://ieeexplore.ieee.org/abstract/document/6512044/}{Less is More:
  Mixed-Initiative Model-Predictive Control with Human Inputs},'' \emph{IEEE
  Transactions on Robotics}, vol.~29, no.~3, pp. 695--703, 2013.

\bibitem{broad2017learning}
A.~Broad, T.~Murphey, and B.~Argall,
  ``\href{http://www.roboticsproceedings.org/rss13/p37.pdf}{Learning Models for
  Shared Control of Human-Machine Systems with Unknown Dynamics},'' in
  \emph{Robotics: Science and Systems}, 2017.

\bibitem{anderson2010optimal}
S.~J. Anderson, S.~C. Peters, T.~E. Pilutti, and K.~Iagnemma,
  ``\href{https://www.inderscienceonline.com/doi/abs/10.1504/IJVAS.2010.035796}{An
  Optimal-Control-Based Framework for Trajectory Planning, Threat Assessment,
  and Semi-Autonomous Control of Passenger Vehicles in Hazard Avoidance
  Scenarios},'' \emph{International Journal of Vehicle Autonomous Systems},
  vol.~8, no. 2-4, pp. 190--216, 2010.

\bibitem{lacey2000context}
G.~Lacey and S.~MacNamara,
  ``\href{http://journals.sagepub.com/doi/pdf/10.1177/02783640022067968}{Context-Aware
  Shared Control of a Robot Mobility Aid for the Elderly Blind},''
  \emph{International Journal of Robotics Research}, vol.~19, no.~11, pp.
  1054--1065, 2000.

\bibitem{argall2009survey}
B.~D. Argall, S.~Chernova, M.~Veloso, and B.~Browning,
  ``\href{http://www.sciencedirect.com/science/article/pii/S0921889008001772}{A
  Survey of Robot Learning from Demonstration},'' \emph{Robotics and Autonomous
  Systems}, vol.~57, no.~5, pp. 469--483, 2009.

\bibitem{levine2013guided}
S.~Levine and V.~Koltun,
  ``\href{http://proceedings.mlr.press/v28/levine13.pdf}{Guided Policy
  Search},'' in \emph{International Conference on Machine Learning}, 2013, pp.
  1--9.

\bibitem{abbeel2004apprenticeship}
P.~Abbeel and A.~Y. Ng,
  ``\href{https://dl.acm.org/citation.cfm?id=1015430}{Apprenticeship learning
  via inverse reinforcement learning},'' in \emph{International Conference on
  Machine learning}.\hskip 1em plus 0.5em minus 0.4em\relax ACM, 2004, p.~1.

\bibitem{ziebart2008maximum}
B.~D. Ziebart, A.~L. Maas, J.~A. Bagnell, and A.~K. Dey,
  ``\href{https://www.aaai.org/Papers/AAAI/2008/AAAI08-227.pdf}{Maximum Entropy
  Inverse Reinforcement Learning},'' in \emph{AAAI Conference on Artificial
  Intelligence}, vol.~8, 2008, pp. 1433--1438.

\bibitem{alshiekh2017safe}
M.~Alshiekh, R.~Bloem, R.~Ehlers, B.~K{\"o}nighofer, S.~Niekum, and U.~Topcu,
  ``\href{https://arxiv.org/abs/1708.08611}{Safe Reinforcement Learning via
  Shielding},'' in \emph{AAAI Conference on Artificial Intelligence}, 2018.

\bibitem{fulton2018safe}
N.~Fulton and A.~Platzer, ``\href{https://nfulton.org/papers/aaai18.pdf}{Safe
  Reinforcement Learning via Formal Methods},'' in \emph{AAAI Conference on
  Artificial Intelligence}, 2018.

\bibitem{reddy2018shared}
S.~Reddy, S.~Levine, and A.~Dragan,
  ``\href{https://arxiv.org/abs/1802.01744}{Shared Autonomy via Deep
  Reinforcement Learning},'' in \emph{Robotics: Science and Systems}, 2018.

\bibitem{brown2018efficient}
D.~S. Brown and S.~Niekum,
  ``\href{https://arxiv.org/pdf/1707.00724.pdf}{Efficient Probabilistic
  Performance Bounds for Inverse Reinforcement Learning},'' in \emph{AAAI
  Conference on Artificial Intelligence}, 2018.

\bibitem{sutherland1968fly}
J.~Sutherland,
  ``\href{http://www.dtic.mil/dtic/tr/fulltext/u2/679158.pdf}{Fly-by-Wire
  Flight Control Systems},'' Air Force Flight Dynamics Lab Wright-Patterson AFB
  OH, Tech. Rep., 1968.

\bibitem{atkeson1997comparison}
C.~G. Atkeson and J.~C. Santamaria,
  ``\href{http://ieeexplore.ieee.org/document/606886/}{A Comparison of Direct
  and Model-Based Reinforcement Learning},'' in \emph{IEEE International
  Conference on Robotics and Automation}, vol.~4, 1997, pp. 3557--3564.

\bibitem{koopman1931hamiltonian}
B.~O. Koopman, ``\href{http://www.pnas.org/content/17/5/315.short}{Hamiltonian
  Systems and Transformation in Hilbert space},'' \emph{Proceedings of the
  National Academy of Sciences}, vol.~17, no.~5, pp. 315--318, 1931.

\bibitem{rawlings2017model}
J.~Rawlings, D.~Mayne, and M.~Diehl,
  \emph{\href{http://jbrwww.che.wisc.edu/home/jbraw/mpc/}{Model Predictive
  Control: Theory, Computation, and Design}}.\hskip 1em plus 0.5em minus
  0.4em\relax Nob Hill Publishing, 2017.

\bibitem{ansari2016sequential}
A.~R. Ansari and T.~D. Murphey,
  ``\href{http://ieeexplore.ieee.org/abstract/document/7562474/}{Sequential
  Action Control: Closed-form Optimal Control for Nonlinear and Nonsmooth
  Systems},'' \emph{IEEE Transactions on Robotics}, vol.~32, no.~5, pp.
  1196--1214, 2016.

\bibitem{abraham2017model}
I.~Abraham, G.~D.~L. Torre, and T.~D. Murphey,
  ``\href{http://www.roboticsproceedings.org/rss13/p52.pdf}{Model-Based Control
  Using Koopman Operators},'' in \emph{Robotics: Science and Systems}, 2017.

\bibitem{tzorakoleftherakis2015controllers}
E.~Tzorakoleftherakis and T.~D. Murphey,
  ``\href{http://ieeexplore.ieee.org/document/7402901/}{Controllers as Filters:
  Noise-Driven Swing-Up Control Based on Maxwells Demon},'' in \emph{IEEE
  Conference on Decision and Control}, 2015.

\bibitem{laskey2017dart}
M.~Laskey, J.~Lee, R.~Fox, A.~Dragan, and K.~Goldberg,
  ``\href{http://proceedings.mlr.press/v78/laskey17a.html}{DART: Noise
  Injection for Robust Imitation Learning},'' in \emph{Conference on Robot
  Learning}, 2017, pp. 143--156.

\bibitem{brockman2016gym}
G.~Brockman, V.~Cheung, L.~Pettersson, J.~Schneider, J.~Schulman, J.~Tang, and
  W.~Zaremba, ``\href{https://arxiv.org/abs/1606.01540}{OpenAI Gym},''
  \emph{arXiv}, vol. abs/1606.01540, 2016.

\bibitem{williams2015data}
M.~O. Williams, I.~G. Kevrekidis, and C.~W. Rowley,
  ``\href{https://link.springer.com/article/10.1007/s00332-015-9258-5}{A
  Data--Driven Approximation of the Koopman Operator: Extending Dynamic Mode
  Decomposition},'' \emph{Journal of Nonlinear Science}, vol.~25, no.~6, pp.
  1307--1346, 2015.

\bibitem{jovanovic2014sparsity}
M.~R. Jovanovi{\'c}, P.~J. Schmid, and J.~W. Nichols,
  ``\href{https://aip.scitation.org/doi/abs/10.1063/1.4863670}{Sparsity-promoting
  Dynamic Mode Decomposition},'' \emph{Physics of Fluids}, vol.~26, no.~2, p.
  024103, 2014.

\bibitem{boyd2004convex}
S.~Boyd and L.~Vandenberghe,
  \emph{\href{http://stanford.edu/~boyd/cvxbook/}{Convex Optimization}}.\hskip
  1em plus 0.5em minus 0.4em\relax Cambridge University Press, 2004.

\bibitem{kalinowska2018online}
A.~Kalinowska, K.~Fitzsimons, J.~Dewald, and T.~D. Murphey,
  ``\href{http://www.roboticsproceedings.org/rss14/p46.pdf}{Online User
  Assessment for Minimal Intervention During Task-Based Robotic Assistance},''
  in \emph{Robotics: Science and Systems}, 2018.

\end{thebibliography}

\end{document}